# Using English as Pivot to Extract Persian-Italian Parallel Sentences from Non-Parallel Corpora

Ebrahim Ansari

Department of Computer Science and Information Technology, Institute for Advanced Studies in Basic Sciences (IASBS), Zanjan, Iran

Department of Computer Science and Engineering, Shiraz University, Shiraz, Iran

M.H. Sadreddini

Department of Computer Science and Engineering, Shiraz University, Shiraz, Iran

Mostafa Sheikhalishahi

Department of Mathematics and Informatics, University of Calabria, Rende, Italy

Richard Wallace

Distributed Systems Architecture Research Group, Complutense University, Madrid, Spain

Fatemeh Alimardani

Department of Computer Science and Engineering, Shiraz University, Shiraz, Iran

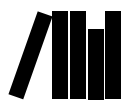

*Ebrahim Ansari, M.H. Sadreddini, Mostafa Sheikhalishahi & Richard Wallace,*
*Fatemeh Alimardani*

The effectiveness of a statistical machine translation system (SMT) is very dependent upon the amount of parallel corpus used in the training phase. For low-resource language pairs there are not enough parallel corpora to build an accurate SMT. In this paper, a novel approach is presented to extract bilingual Persian-Italian parallel sentences from a non-parallel (comparable) corpus. In this study, English is used as the pivot language to compute the matching scores between source and target sentences and candidate selection phase. Additionally, a new monolingual sentence similarity metric, Normalized Google Distance (NGD) is proposed to improve the matching process. Moreover, some extensions of the baseline system are applied to improve the quality of extracted sentences measured with BLEU. Experimental results show that using the new pivot based extraction can increase the quality of bilingual corpus significantly and consequently improves the performance of the Persian-Italian SMT system.

# 1 Introduction

In recent years, machine translation (MT) systems have obtained reasonable results when applied to some popular language pairs such as English-French and English-Chinese. Over the last few decades many different approaches for MT have been proposed, such as rule-based approaches (direct translation, interlingua-based, transfer-based) and corpus-based (statistical, example-based) and hybrid approaches which use both former approaches together. However, studies on statistical MT for low-resourced languages are always faced with the challenge of getting enough data to support any particular approach. Statistical machine translation (SMT) uses statistical methods based on large parallel bilingual corpora of source and target languages to build a statistical translation model. SMT also uses target language texts to build a statistical language model. These two models and a search (decoding) module are used to decode and find the best translation for each source language sentence (Brown et al. 1993; Koehn, Och & Marcu 2003) . The performance of the system heavily depends on the quality and quantity of the corpus used for training. For some language pairs, such as Persian-Italian, there aren't enough parallel corpora for the training phase that eventually builds a statistical machine translation system. In recent years, most of the proposed methods to solve this problem are based on two well-known schemas; (a) using pivot languages which is described in Section 1.1 and (b) extracting parallel sentences from non-parallel bilingual corpora described in Section 1.2.

This paper describes a new approach for using comparable corpora to extract Persian-Italian parallel corpus and eventually improve SMT system performance for this under-resource language pair. In our proposed method, instead of using a source-target machine translation system to extract parallel sentences, similar sentences are found with comparison of source-pivot and target-pivot translations with considering similarity metrics. In our experiments, English language is used as the pivot.

## 1.1 Using a Pivot Language

Recent research on multilingual SMT focuses on the use of pivot languages as a solution to overcome language resource limitations (Gispert & Mario 2006; Utiyama & Isahara 2007; Wu & Wang 2007; Bertoldi et al. 2008; Philipp, Alexandra & Ralf 2009; Paul et al. 2011; Nakov & Ng 2012). Pivot based machine translation refers to the use of an intermediate language, called pivot language, to translate from the source language to the target language. Unlike typical MT systems, which translate directly from source language to target language, pivot based systems translate sequentially from source to pivot and then from pivot to target language. (Utiyama & Isahara 2007) define two famous pivot strategies; i.e. Sentence translation and phrase translation. The sentence translation strategy means that a French sentence is translated into several English sentences, and then subsequently all translations should be translated into German separately. Then, we select the highest scoring sentence from the German sentences. The phrase translation strategy means that we directly construct a French-German phrase translation table from combining a French-English phrase-table and an English-German phrase-table. We assume that these French-English and English-German tables are built using the phrase model training in the baseline statistical machine translation system as just described in (Philipp, Alexandra & Ralf 2009), which means phrases are heuristically extracted from word-level alignments produced by doing training on the corresponding parallel corpora (Utiyama & Isahara 2007).

Another way of using pivot language is to use the existing source-pivot and pivot-target machine statistical systems to create new source-target bilingual and parallel corpus which have some similarity with the former method. However, instead of creating source-target phrase translation, the new source-target corpora are created using translation tables in primary corpora.

Moreover, (Nakov & Ng 2012) use a third language, not literally as a pivot language, to improve their machine translation system. Their method is applicable when first, there is a small parallel corpus between source-target languages and there is sufficient parallel corpus between target-third languages and then, source and third languages are similar. The improvement is achieved by taking advantage of the vocabulary overlap and similarities between the source and

*Ebrahim Ansari, M.H. Sadreddini, Mostafa Sheikhalishahi & Richard Wallace, Fatemeh Alimardani*

third languages in spelling, word order, and syntax.

The usage of pivot language in our work is slightly different from previous strategies because we used it as a part of our main implementation. In our proposed work, the English language is used as the pivot to translate Persian and Italian parts of selected sentences of comparable Persian-Italian corpora to one similar language (English) to eventually find the best matches and to extract final parallel sentences. Two different and pre-created Persian-English and Italian-English translation systems are used to translate source and target parts to English.

### 1.2 Using Comparable Corpora

Comparable corpus as a source of translation knowledge has attracted the attention of many researchers. Unlike parallel corpora, which are clearly defined as translated texts, there is a wide variation of non-parallelism in comparable texts. Non-parallelism is manifested in terms of differences in author, domain, topics, time period and language. Most common text corpora have non-parallelism in all these dimensions. The higher the degree of non-parallelism, the more challenging is the extraction of bilingual information (Chakraborty 2012). Words, phrases, clauses, sentences, and discourses contain various levels of parallelism, due to word-to-word translation depending on the corpora characteristics, but compared with true parallel corpora, are still not considered parallel. Comparable corpora are much easier to build from commonly available documents, such as news article pairs describing the same event in different languages.

Recently, there has been increased interest in the automatic creation of parallel corpora. Apart from alignment methods which extract parallel sentences from two monolingual corpora (Parallel corpus) (Koehn 2005; Gale & Church 1993) several researchers (Resnik & Smith 2003; Kilgarriff & Grefenstette 2003; Utiyama & Isahara 2003; Yang & Li 2003; Fung & Cheung 2004; Munteanu & Marcu 2005; 2006; Kumano, Tanaka & Tokunaga 2007; Abdul-Rauf & Schwenk 2009; Rauf & Schwenk 2011; Diep, Besacier & Castelli 2010; Fung, Prochasson & Shi 2010; Smith, Quirk & Toutanova 2010; Hewavitharana & Vogel 2011) have shown that fairly good quality parallel sentence pairs can be automatically extracted from parallel and comparable corpora, which subsequently could be used to improve the performance of machine translation (MT) systems. In Section 2.1 of this paper, there is a brief review of this research.

This paper proposes a new approach to extract Persian-Italian parallel texts from comparable corpora. The proposed approach has some similarity to (Rauf & Schwenk 2011)'s work. In our version, instead of finding the best matches

*1 Using English as Pivot to Extract Persian-Italian Parallel Sentences from Non-Parallel Corpora*

between source-target translations and target part of comparable corpora, the similar sentences are found with comparison of source-pivot and target-pivot translations. Furthermore, some new mechanisms and an additive similarity metric have been added to the previous work's baseline. In our experiments, English language is used as the pivot.

While our translated parts are not human written sentences and are translated with machine, and the chance of having good syntax structure is low, using a semantic similarity metric is more reliable and useful. In recent years, a few methods have been proposed to compute sentence similarities (Mandreoli, Martoglia & Tiberio 2002; Hatzivassiloglou, Klavans & Eskin 1999; Mihalcea, Corley & Strapparava 2006; Li et al. 2006; Liu, Zhou & Zheng 2007; Xiao-Ying, Yi-Ming & Ruo-Shi 2002; Achananuparp, Hu & Shen 2008; LI et al. 2009; Aliguliyev 2009).

(Aliguliyev 2009) uses a method to measure dissimilarity between sentences using the Normalized Google Distance (NGD) (Cilibrasi & Vitanyi 2007). The NGD takes advantage of the number of hits returned by Google to compute the semantic distance between concepts. The concepts are represented with their labels which are fed to the Google search engine as search terms. In our approach, to modify the Information Retrieval (IR) system which is used before in classic framework (Rauf & Schwenk 2011), another sentence similarity metric is used and has been added to our system. This metric is very similar as the score introduced by (Aliguliyev 2009) and we combine it with our existed IR system in the way that we first perform one primary similarity check and then add it to scoring mechanism of existed IR system to calculate similarities between sentences. We show that this metric could improve the extracted sentences in comparison with our initial method. In Section 2, we have reviewed some related works to our implementation, and in Section 3 our work has been described. Section 4 explains the experimental results and finally our conclusion is proposed in Section 5.

## 2 Related Works

In this Section related works to our approach are reviewed. In Section 2.1 works in comparable corpora domain are reviewed and in Section 2.2 works related to our similarity metric are reviewed.

### 2.1 Comparable Corpora

One way to solve the lack of parallel data problem is to exploit a much more available and diverse resource such as comparable non-parallel corpora. Compa-rable corpora are texts that, while not parallel in the strict sense, are somewhat related and convey overlapping information. For example, multilingual news feeds produced by news agencies and different versions of the same articles in Wikipedia could be treated as comparable corpora. There has been a considerable amount of work on bilingual corpora to extract parallel sentences. These include treating the web as a source of semi-parallel sentences (Fung, Prochasson & Shi 2010; Resnik & Smith 2003) or using other sources of comparable corpora such as extracted documents from different news agencies.

(Resnik & Smith 2003) uses their structural filtering system STRAND which filters candidate parallel pairs by determining a set of pair-specific structural values from the underlying HTML page. They report a precision of 98% and a recall of 61% on their developed English-Chinese parallel corpus.

(Zhao & Vogel 2002) proposes a maximum likelihood criterion which combines sentence length model and a statistical translation lexicon model extracted from an already existing aligned parallel corpus. An iterative process was applied to retrain the translation lexicon model with the extracted data. Their selected languages were Chinese and English.

(Utiyama & Isahara 2003) use cross language information retrieval techniques and dynamic programming to extract parallel sentences from an English-Japanese news corpus. The authors first try to find similar article pairs, and then, they treat these pairs as parallel texts, align their sentences on a sentence pair similarity score. Subsequently, they use dynamic programming to find the minimum-cost alignment over each document pair. They use the BM25 similarity measure for their algorithm. (Yang & Li 2003) proposed a parallel sentences extraction schema to identify Chinese and English title pairs based on dynamic programming.

(Fung & Cheung 2004) present a method to extract parallel sentences from very non-parallel corpora by exploiting bootstrapping on top of IBM Model 4 (Brown et al. 1993). They claim that their "find-one-get-more" strategy principle allows them to add more parallel sentences from dissimilar documents, to the baseline set. Primary steps of their method is alike the former approaches while they uses similarity metric same other approaches. Then they used an iterative bootstrapping framework based on the principle of "find-one-get-more", which claims that documents found to contain one pair of parallel sentences must contain others even if the documents are judged to be of low similarity. They rematch documents with using extracted sentence pairs, and refine the mining process iteratively until convergence.

(Munteanu & Marcu 2005), first used a dictionary to translate some of the words of the source sentences, and then used these translations to query a database for finding matching translation candidates and extracting final parallel sentences. In other work, (Munteanu & Marcu 2006) train a maximum entropy classifier to extract parallel corpus in Arabic, English and French languages. They show that a good-quality MT system can be built from scratch by starting with a very small parallel corpus (100,000 words) and exploiting a large non-parallel corpus.

(Kumano, Tanaka & Tokunaga 2007), proposed a method of extracting phrasal alignments from comparable corpora by using an extended phrase-based joint probability model for statistical machine translation (SMT). They have indicated that their method does not require preexisting dictionaries or splitting documents into sentences in advance.

(Abdul-Rauf & Schwenk 2009) present another technique similar to (Munteanu & Marcu 2005)'s and use a statistical machine translation system instead of the bilingual dictionary. In their approach they used an IR system to find the best candidates from translated sentences. Moreover, they used well-known evaluation metrics WER (Word Error rate), TER (Translation Error Rate) and TERp (Translation Error Rate plus) to decide the degree of parallelism between candidate sentences. (Diep, Besacier & Castelli 2010) presents an unsupervised method and use statistical translation system to detect parallel French-Vietnamese parallel sentences with mining comparable corpora. An iterative process was implemented to increase the number of extracted parallel sentence pairs which improved the overall quality of the translation.

Some other approaches are focused on parallel corpus production to improve SMT systems (Fung, Prochasson & Shi 2010; Smith, Quirk & Toutanova 2010; Hewavitharana & Vogel 2011; Ion 2012). (Fung, Prochasson & Shi 2010) use the web as a comparable resource to extract potential parallel sentences with using web crawler. They propose a sentence extraction architecture inspired by various pieces of earlier works. They've reported interesting results using French and English Wikipedia articles as their comparable resource. (Smith, Quirk & Toutanova 2010) propose a parallel sentences extraction method by modeling the document level alignment. Their work is motivated by the observation that parallel sentence pairs are often found in close proximity. They used the same documents in different languages in Wikipedia as their document level aligned corpora. In addition, they used some features which created by using the additional annotation given by Wikipedia, and features made by an automatically induced lexicon model. (Hewavitharana & Vogel 2011) used and evaluated three phrase alignment approaches to detect parallel phrase pairs embedded in comparable sentences: First approach relies on the Viterbi path, second one does not rely on the Viterbi path and uses only lexical features and the third approach is a binary classifier that detects parallel phrase pairs when presented with a large collection of phrase pair candidates. They've claimed that their Non-Viterbi alignment approach outperforms the other two approaches on F-1 measure. F-1 measure is generated from combination of precision and recall measures and is:

$$F - 1 = \frac{2 * Percision * Recall}{Percision + Recall} \qquad (1.1)$$

(Ion 2012) has implemented the freely available toolkit, PEXACC[1], a distributed trainable parallel sentence/phrase extractor from comparable corpora. Moreover, he shows that the performance of a parallel sentence extractor depends on the degree of comparability.

The effect of comparability degree on the accuracy of extracted data is another issue which is considered in recent years. (Li & Gaussier 2010) (Li, Gaussier & Aizawa 2011) have introduced a new measure to calculate and evaluate the degree of comparability of a given comparable corpus. However they've used this measure to improve their comparable corpora and extract parallel lexicon instead of parallel sentences. (Li & Gaussier 2010) propose a measure to calculate the comparability degree and use an iterative construction process to improve the quality of a given comparable corpus to extract bilingual lexicon. (Li, Gaussier & Aizawa 2011) use a cluster based method to enhance corpus comparability based on the measure they introduced before.

The baseline of our proposed approach is similar to that of (Rauf & Schwenk 2011) though we added the idea of pivot language. The reason of using pivot based mechanism is because of the shortage in parallel Persian-Italian resources. (Diep, Besacier & Castelli 2010) use a pivot based mechanism in their work. Other ideas such as using additive similarity metric in IR process and using inverted translation in filtering phase are added to classic approach indeed.

## 2.2 Similarity metrics

Similarity measures and techniques play an important role in Natural language Processing (NLP) and IR applications. There is a large amount of literature on measuring the similarity between documents or long texts (Meadow, Boyce & Kraft 2000), but there are a limited number of research relating to similarity calculation between sentences. Famous techniques for calculating the similarity level between documents usually consider and count shared words between documents (Gerald 1988). Such methods are not suitable to measure the similarity between sentences because the similar long texts usually contain a degree of co-occurring words, while in similar sentences, word co-occurrence may be very rare. In classic approaches for calculating sentence similarity, semantic and syntactic information should be considered together, because both of them contribute to the meaning of the overall sentence (Liu, Zhou & Zheng 2007).

---

[1] Parallel phrase Extractor from Comparable Corpora

(Mandreoli, Martoglia & Tiberio 2002) proposed a method named "approximate sub2 sequence matching" which adopts the Edit Distance as similarity measure between parts of sentences. Their time consuming approach is purely syntactic and focuses on the similarity of syntax structure. (Hatzivassiloglou, Klavans & Eskin 1999) presented a composite similarity metric over small textual units, which utilize semantic information. They have used machine learning technique to select optimal combination between several potential features. (Mihalcea, Corley & Strapparava 2006) present another method for measuring the semantic similarity of short texts. They used some corpus and knowledge based similarity measures that can be calculated using the similarity of the component words.

(Li et al. 2006) present an algorithm that takes account of semantic information and word ordering together to calculate similarity between very short texts (i.e. sentences). They used information from lexical database and corpus statistics to calculate the semantic similarity of two sentences. (Liu, Zhou & Zheng 2007) propose another method to calculate sentence similarity with considering semantic information and word ordering. ("Measuring semantic similarity within sentences") present a novel method which in addition of semantic information and word orders, considers the contribution of different parts of speech (POS) in a sentence. They use the analysis of this POS to calculate and measure the semantic similarity by a Dynamic Time Warping (DTW) technique.(Achananuparp, Hu & Shen 2008) tested different combinations of sentence vector similarity, word order similarity, POS similarity and also took into account question category similarity to measure the question similarities. They have evaluated different similarity measures on three different dataset. Their work could be used as a survey to see the effect of different similarity measures. (LI et al. 2009) measure and evaluate sentence similarity from different aspects. They define Objects-Specified Similarity, Objects-Property Similarity, Objects-Behavior Similarity and Overall Similarity to determine sentence similarities from four aspects. They claim that these similarities are based on the information people get from a sentence.

(Aliguliyev 2009) uses a new sentence similarity metric for their text summarization method. Their presented approach uses NGD (Cilibrasi & Vitanyi 2007)

*Ebrahim Ansari, M.H. Sadreddini, Mostafa Sheikhalishahi & Richard Wallace, Fatemeh Alimardani*

to measure dissimilarity between sentences. This metrics is based on semantic based similarity calculation and does not consider syntax criteria. In our implementation, whereas two sides of sentences are translated with other statistical machine translation systems and not by human, the translation may be not very accurate and the syntax based similarity metrics are not very useful. Therefore, the authors decide to use the idea presented in (Aliguliyev 2009) to calculate sentences similarity of source to pivot and target to pivot parts. NGD uses the number of hits returned by Google to compute the semantic distance between concepts (terms). The concepts are represented with their labels which are fed to the Google search engine as search terms. In (Aliguliyev 2009) a dissimilarity concept is used instead of similarity. The reason of this choice is based on the point that the NGD is nonnegative and does not satisfy the triangle inequality, i.e. they shall name it dissimilarity measure not similarity. Using the NGD, the (Aliguliyev 2009) defines the global and local dissimilarity measure between terms. In our implementation, the global dissimilarity measure is used as a part of our similarity score and the local score is neglected. Formula 1:

$$NGD(t_i, t_j) = \frac{\max \log(f(t_i)), \log(f(t_j)) - \log(f(t_j, t_j))}{\log(N) - \min \log(f(t_i)), \log(f(t_j))} \quad (1.2)$$

Where $f(t_i)$ is the number of pages that contain the search term $t_i$ and $f(t_j, t_j)$ denotes the number of web pages containing both terms $t_i$ and $t_j$ together. $N$ is the number of web pages indexed by Google. As described in (Cilibrasi et al. 2007), the range of the NGD is between 0 and $\infty$: If $t_i = t_j$ or if $t_i \neq t_j$ but $f(t_i) = f(t_j) = f(t_i, t_j) > 0$, then $NGD(t_i, t_j) = 0$; this means that the semantics of $t_i$ and $t_j$, in the Google sense is the same. If $f(t_i) = 0$, then for every term $t_i$, we have $f(t_j, t_j) = 0$, and the $NGD(t_i, t_j) = \frac{\infty}{\infty}$, which we define it to be 1. If $f(t_j) \neq 0$ and $f(t_j, t_j) = 0$, we take $f(t_j, t_j) = 1$.

Finally, and with using the formula 1 the global dissimilarity measure between sentences $s_a$ and $s_b$ are defined as follows: Formula 2:

$$DisNGD(s_a, s_b) = \frac{\sum_{t_i \in s_a} \sum_{t_j \in s_b} NGD(t_i, t_j)}{m_i \dot{m}_j} \quad (1.3)$$

In (Aliguliyev 2009) this is the part of overall dissimilarity measure and they named it the global part. This score could be used as a sub-score measure. In our implementation we added the measure calculated from Formula 2 to the score calculated from our IR system to have a modified scoring system. Overall scores are used to find candidate parallel sentences for further checking.

## 3 Our Approach

In our approach we extract Persian-Italian corpus from non-parallel corpora based on the mechanism which is described in this section. Figure 1 shows a brief overview of our implementation. We start by translating both source and target parts of our corpora to the pivot-language. After the translation phase there are two English corpora which were named EnPr (English translated from Persian) and EnIt (English translated from Italian). The first one is translated sentences from Persian and the second is a translation from Italian as part of our primary comparable corpora. In this step, if we could find similar sentences in both English parts, consequently, corresponding Persian and Italian sentences are probable translations of each other. To find similar sentences between two translated English parts EnPr and EnIt, the IR system that (Rauf & Schwenk 2011) introduced can be used. Although they used IR system to find candidate sentences between source to target translation and target part, in our work we have used the IR for the two translation parts. Moreover, we have used another similarity metrics (NGD) which are more semantic-based.

After finding candidate sentences, i.e. one EnPr sentence and top n candidate sentences from EnIt, these candidates are compared to find the best pairs and consequently to add the reasonable ones to our parallel Persian-Italian corpus. For this purpose, similarly to (Rauf & Schwenk 2011)'s work, candidates are filtered out by using simple filters like WER, TER and TERp. In our approach we used and examined another strategy in the filtering phase. In this new strategy, we first select the corresponding Persian sentences for EnPr part of candidates, and translate them into Italian using a small Persian-Italian machine translation system and these new inverted translation sentences are sent to filtering phase with Italian part of extracted candidates.

(Rauf & Schwenk 2011) have shown that quality of the SMT system does not seem to affect the IR process. Thus, a fairly simple SMT system built on small amounts of parallel text can be used to extract good parallel sentences from a comparable corpus. Consequently, the resources required by our system are as the following:

- 1) A Source to pivot parallel text to create a translation model of the first SMT system

- 2) A Target to pivot parallel text to create a translation model of the second SMT system

- 3) Monolingual data in the pivot language to train the language model of both SMT systems

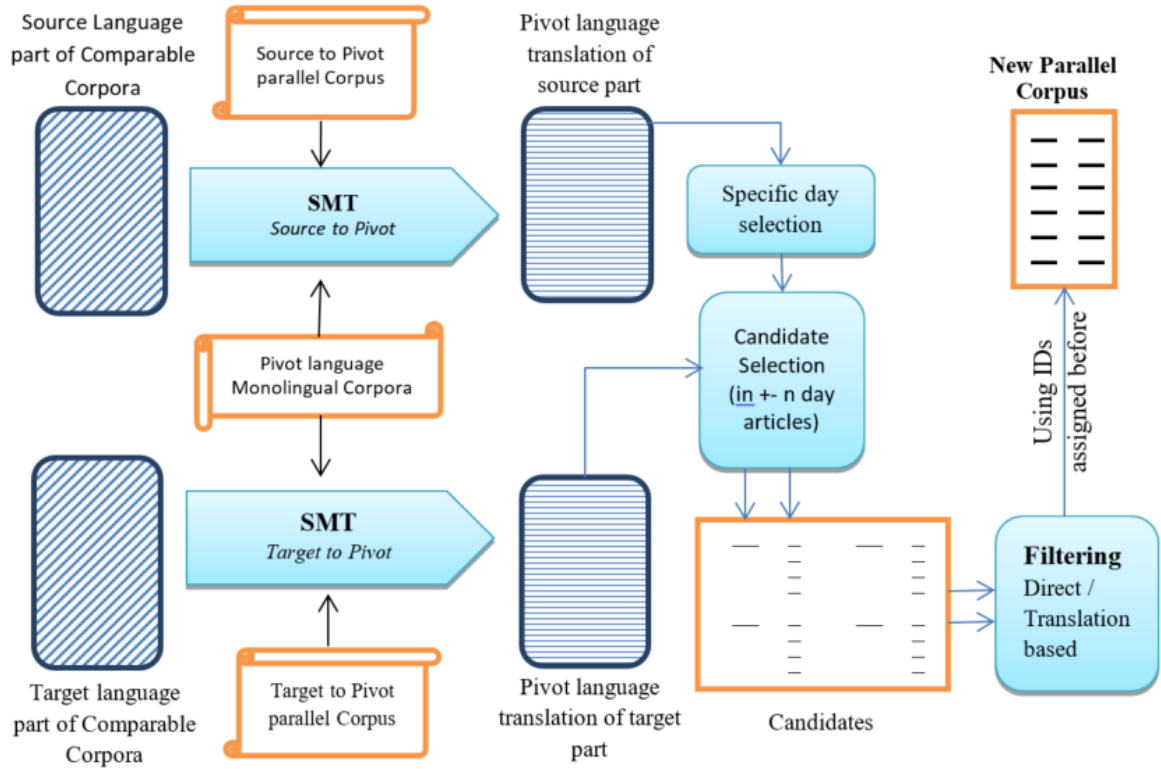

Figure 1: Overview of the parallel sentence extraction system. Both source-language (Persian in our work) and target-language (Italian) sides of the comparable corpus are translated into the pivot-language (English in our case)

### 3.1 Translating both parts to English

Two different statistical machine translation systems are needed to translate the source and the target parts into one pivot language. The translation modules in this work are built using Moses toolkit (Koehn et al. 2007) with the default setting and as follows:

- GIZA++ ((Och & Ney 2003)) was used for word alignments, the "-alignment" option for phrase extraction was "grow-diag-final-and"

- Fourteen features in total were used in the log-linear model: distortion probabilities (six features), one 3-gram language model probability, bidirec-tional translation probabilities (two features) and lexicon weights (two features), a phrase penalty, a word penalty and a distortion distance penalty.

- A 3-gram English language model was built using the SRILM toolkit with using our monolingual English corpus for Persian-English and Italian-English SMTs and a 4-gram Italian language model that was built in the same way with using our large monolingual Italian corpus.

By using the above setting, two Persian-English and Italian-English MT systems were built to translate Persian and Italian sentences to English. While the amount of bilingual corpora for Persian-English and Italian-English is not very low, these two systems provide good translations in comparison with direct Persian-Italian translation system.

## 3.2 Candidate selection

In this phase candidate sentences from both translated parts are extracted and consequently are sent to the filtering phase for final checking. As described in Section 2.1, (Rauf & Schwenk 2011) used a specific-day based IR system to find candidates. Their sentence extraction process is based on a simple and apparent heuristic, which assumes that a news item reported on day X in one news corpus will be most probably found in a 5-day time window in another corpus. In our work an extra similarity metric, Normalized Google Distance, is introduced in addition to previous work.

### 3.2.1 Using IR system

Similar to (Rauf & Schwenk 2011) 's work we have used the Lemur toolkit [2] (Ogilvie & Callan 2001) as our IR system. The day-specific filter is used with consideration of using the ID and date information for each sentence of both corpora. After the translation phase we have two English corpora; one from Persian and one from Italian. The goal of IR system is to find best candidate sentences in EnIt (English translated from Italian part) for each selected EnPr (English translated from Italian part) sentence.

As the first step, all sentences from the EnPr for a specific day (query sentences) and the corresponding articles from EnIt in a multiple day window are collected.

---

[2] The Lemur Toolkit is designed to facilitate research in language modeling and information retrieval (IR), where IR is broadly interpreted to include such technologies as ad hoc and distributed retrieval with structured queries, cross-language IR, summarization, filtering, and categorization.

*Ebrahim Ansari, M.H. Sadreddini, Mostafa Sheikhalishahi & Richard Wallace,*
*Fatemeh Alimardani*

The Lemur IR toolkit is used for sentence extraction. IR phase is the most time consuming part of the system. In order to reduce it, using limited window day such as $\pm 5$ days instead of larger windows, or having less candidates output seems to be obligatory. In order to decrease the length of query sentences, and consequently to increase the efficiency of IR toolkit, stop words are removed from sentences. The top five scoring sentences are returned by the IR process in this phase. Consider that these top scoring sentences are in English language. Therefore for next phases and eventually for the parallel sentences extraction step, the original sentences (i.e. Persian and Italian ones) are needed. With using a defined ID in our implementation, corresponding Persian and Italian sentences are selected from EnPr and EnIt and will be send to next phase.

### 3.2.2 Modified IR with the new similarity measure

We use another candidate selection metric, Normalized Google Distance (NGD), in addition to what the (Rauf & Schwenk 2011) used. This similarity metric could calculate the similarity between each two sentences in one language and is very similar to the global version that was introduced in (Aliguliyev 2009) and is fully described in Section 2.2. We have used our similarity metric in the IR system in two ways: First for filtering sentences from IR search space to cut some process time; and then combining the scores with the Lemur scoring system to increase the efficiency of candidate selection.

The IR process for each query sentence is very time consuming. Because of this problem we must use a small time window or to decrease the number of extracted candidates for reasonable process time. It is obvious that if we could decrease the search space for each query sentence, the IR process could be done faster and consequently we could use our IR system with bigger time window or more candidate extraction setting. Therefore, for each query sentence in the left side (EnPr translated part), the similarity scores with all of sentences in the right side (selected EnIt articles) are calculated first and top X sentences are selected as the IR system search space. In our setting, the value of X is 40%. Based on our experiments the increasing of the value of the X threshold does not have any significant improvement in our extracted data.

Another usage of the new scoring metric in our system is to combine it with our IR system scoring method. For this purpose, we changed the described IR system in (Rauf & Schwenk 2011) to use these pre-calculated scores (which were calculated in the previous phase for all sentences pairs) as one additive value in the IR system candidate selection and its indexing process phase. We named this new IR system the Modified IR. With this improved IR system, the number of selected sentences could be as much as 10 and the time window is increased to 7 days. This point must be considered that in our settings, in addition of the IR system output, top N sentences with least NGD (i.e. most similarity score), will be selected as candidates for next phase. Therefore, the candidates (selected sentences) can be defined as:

$$Candidates = Top(X)_{NGD} \cap Top(M)_{ModifiedIR} \qquad (1.4)$$

$$Candidates = Top(N)_{NGD} \cup Top(M)_{ModifiedIR} \qquad (1.5)$$

where in our experiments, X is 40%, N is equal to 5 and M is 7.

## 3.3 Filtering phase

The final phase of sentence extraction system is to select parallel sentences from candidates. This task could be done with a simple similarity metrics as was done in (Rauf & Schwenk 2011). Moreover, in our work an extra filtering method called inverted candidate translation strategy is used. In the following sub-sections, the normal and the new extended versions are reviewed respectively. We have used tail removal procedure as introduced in (Rauf & Schwenk 2011) in the final step of our work.

### 3.3.1 Filtering based on candidate sentences

In this step the sentences with best matching scores are selected among candidate sentences obtained from previous part. The inputs of this step are English translated sentences from Persian and Italian corpora. The original sentences from the Persian/Italian corpora are now retrieved by using the pre-defined ID and are considered as potential parallel translation. Even though Abdul-Rauf and Schwenk (2011) declared that TER-selected sentences score slightly better than those selected by WER and TERp, but because of differences between our work and theirs, we examined the three similarity metrics WER, TER and TERp individually in our work. WER measures the number of operations required to transform one sentence into the other (by counting the number of insertions, deletions and substitutions). This is obvious that there are many correct but different translations for any given sentence, each of them might be different not only in the word choice but also in word ordering which WER couldn't take it into account. This problem is addressed by TER which allows block movements of words, and therefore, could consider the reordering of words and phrases in

*Ebrahim Ansari, M.H. Sadreddini, Mostafa Sheikhalishahi & Richard Wallace, Fatemeh Alimardani*

translation (Snover et al. 2006). TERp (Snover et al. 2009) is an extension of TER and was one of the top-performing metrics at the NIST Metric MATR workshop [3] and it had the highest absolute correlation, as measured by the Pearson correlation coefficient, with human judgments in nine of the forty five test conditions. TERp uses all the edit operations of TER (matches, insertions, deletions, substitutions and shifts) as well as three new edit operations: stem matches, synonym matches and phrase substitutions. In general, TERp allows better sentence comparison by incorporation of a kind of linguistic information about words. The introduced filters inherently impose a penalty for large difference between lengths of two sentences, therefore the sentence-length filtering is not used in our work.

### 3.3.2 Filtering based on source to target translation

In this section, we introduce another filtering method which is different from previous approaches. In this method, instead of checking translated parts (EnPr and EnIt) with each other to find the best sentence between candidates, the source to target translation is used for further checking. For this purpose, the third machine translation system, Persian-Italian SMT, is introduced. The existing training data for this machine translation system is not very large, therefore this system can be just used for this filtering part.

First of all, by using the ID defined for sentences in EnPr and EnIt (two translated corpora) the original sentences are retrieved from Persian and Italian primary corpora and will be considered for potential best sentence pair selection. Consequently, there are one Persian sentence and N Italian candidates. Using Persian-Italian created SMT, The Persian sentence is translated to Italian and by using one of the selected sentence similarity metrics (i.e. WER, TER or TERp) the best candidate from Italian sentences will be selected as parallel translation.

Even though (Rauf & Schwenk 2011) declared that the amount of parallel data does not have a big effect on the system, a better source to target machine translation system (e.g. having bigger parallel corpora) could lead to better filtering and consequently better sentence selection. In our work, the idea of using top translated sentences is used instead of using just one translated sentence. Our original Persian-Italian SMT system first generates the n-best Italian translations and then it selects the best of them. In our extension the top n best translations (in our work n is equal to 5) will be considered as the output of the SMT. The similarity scores of all these top five Italian sentences with each Italian candidate

---

[3] http://www.nist.gov/itl/iad/mig/metricsmatr10.cfm, Metrics for Machine Translation 2010 Evaluation(MetricsMaTr10)

are calculated and their sum is considered as the filtering score. For example if sentence similarity score is TER, this new score is calculated as follows:

$$TER_{total}(i) = \sum_{j=1}^{n} TER(translation_j, candidate_i) \qquad (1.6)$$

In experimental results, the effect of using this inverted translation scoring are evaluated.

## 4 Experimentation

In this section, the properties and effects of our parallel extraction system will be evaluated. First, the comparable corpora and the machine translation systems used in this work are introduced. Section 4-2 presents the properties of extracted data with and without used similarity metric in IR system, the effects of using different filtering strategies, and finally the effects of adding different extracted corpora to the training data of baseline MT systems.

### 4.1 Used data and SMTs

To obtain our comparable corpora, we collected news from different Persian and Italian agencies between years 2009-2012. We divided our data (Both Persian and Italian) to two different classes. First all the news related to international sports news and second was all international news collected from news agencies. It is obvious that the former corpus is much comparable in comparison with latter one. All the documents are indexed with date of publishing. The Persian part of corpora contains: International sport related news from ISNA[4], Fars news agency [5], and Persian part of Goal.com [6] website. International news collected from ISNA [7] and Fars news agency [8]. Our Italian part was selected from these resources: International sport related news from CORRIERE DELLA SERA[9] and

---

[4] ISNA, Iranian students News Agency, Sport News part, Persian, http://isna.ir/fa/service/Sports
[5] Fars News Agency, Sport News part, Persian, http://www.farsnews.com/newsv.php?srv=4
[6] Goal.com, Persian part, http://www.goal.com/iran
[7] Fars News Agency, International News part, Persian, http://www.farsnews.com/newsv.php?srv=6
[8] ISNA, Iranian students News Agency, International News part, Persian, http://isna.ir/fa/service/World
[9] CORRIERE DELLA SERA, Sport News part, Italian, http://www.corriere.it/sport/

*Ebrahim Ansari, M.H. Sadreddini, Mostafa Sheikhalishahi & Richard Wallace, Fatemeh Alimardani*

international news from La Gazzetta dello Sport [10] International news collected from La Repubblica [11] and CORRIERE DELLA SERA [12],

Table 1 shows the properties of initial comparable corpora used in our experiments. As introduced in Section 3, two Persian-English and Italian-English parallel corpora are used to create two translation models of primary statistical machine translation systems. With using these systems different parts of our comparable corpora are translated to pivot language, English, and will be sent for further processes. The sizes of Persian-English and Italian-English parallel corpora were about 830 thousand and 1.920 million Sentences respectively. In addition to above corpora, in order to improve our language models, two source-target dictionaries are used. The Persian-English dictionary has 200 thousand entries and the second one, Italian-English dictionary's size is about 300K. Finally, we use one 4-gram model (created from 100M words mono-lingual English texts) to train language model of both systems.

Table 1: The properties of used comparable Corpora

| Name | Number of articles (K) | Number of words (M) |
|---|---|---|
| *Persian Sport news* | 98 | 15 |
| *Italian Sport news* | 115 | 18 |
| *Persian general news* | 494 | 107 |
| *Italian general news* | 624 | 151 |

Based on the idea of using a source to target SMT, another parallel corpus (a small Persian-Italian corpus) was needed. This corpus is collected from subtitles[13] and a Wikipedia generated Corpora using our Wikiretriever[14] system. One Italian text is used for training our 4-gram language model. All of our the three needed machine translation systems (Persian-English, Italian-English and Persian-Italian) are created by using Moses SMT toolkit (Koehn et al. 2007) with the default setting as described in Section 3.1.

---

[10] La Gazzettadello Sport, Italian, http://www.gazzetta.it/

[11] CORRIERE DELLA SERA, International news, Italian, http://www.repubblica.it/

[12] La Repubblica, Italian, http://www.repubblica.it/

[13] Retrieved from OPUS (the open parallel corpus) http://opus.lingfil.uu.se/. OpenSubtitles2011 and OpenSubtitles2012 parallel corpora

[14] Wikiretriever is a C++ crawler to retrieve very accurate parallel sentences from two different languages of Wikipedia website and has written to retrieve limited amount of parallel sentences by using dates, numbers and famous words. The accurate of this retriever for English and Persian which has been evaluated before is about 98%, so is very ideal for our purpose.

## 4.2 Extracted data

In our experiments we used several and different settings to evaluate our proposed ideas. In all test cases, twenty percent of data was selected as the development data and the rest was selected as test data and we have added top 50% of extracted sentences to our small existing parallel corpus. In the first step, we decided to check the effects of adding the recent similarity metric (Normalized Google Distance) in two points: 1) at the primary filtering phase and 2) on the IR system. For this evaluation, we applied four different schemas. First, we used the simple pivot based system without NGD. The second schema was the system which utilizes the NGD distance in its primary phase and not in the IR system. In third and fourth schemas we used our modified IR system with and without the NGD in their primary phase. As described before, for all tests with NGD in their primary phase, we have taken advantage of the NGD's higher pruning rate by setting the time windows to 7. For all other tests (without NGD in their primary phase) time window is set to five. Moreover, in order to check the effect of using source to target translation in filtering phase, we did our experiments with and without our proposed inverted translation idea. The results of these experiments are shown in Figure 2 and Figure 3. A general comparable corpus is used in Figure 2 and international sport based texts are considered as the comparable corpus in Figure 3. In all of them, the system uses pivot based schema (Persian-English and Italian-English SMTs) to create second layer comparable data. Consequently, in the filtering phase in order to find the best extracted parallel sentence, TER filter is selected. In the legends of these Figures, "google" means using Normalized Google Distance in preliminary phase, "ModifiedIR" means using Modified IR version (i.e. using Normalized Google Distance in our IR system). Figure 2 (as well as Figure 3) shows the positive effect of using the similarity metric (Normalized Google Distance) in both preliminary phase and modified IR system. So these new modifications could improve our initial extracted sentences. As can be seen in this figure, using inverted translation for candidates, in some cases but not always, give us slightly better efficiency. As indicated before, the amount of existing parallel corpus to build Persian-Italian SMT is very limited. The authors believe that if the amount of this accurate corpus was sufficient, using inverted translation could improve the results. Another issue which should be considered is how the comparability degree will affect the final results. In our experiments, as indicated in Figures 2, and 3, to build primary SMTs, two different corpora were used: a domain based version and a general version. The domain based corpus contains the international sport related texts and because of similarity between these types of news in different languages, it is more comparable.

*Ebrahim Ansari, M.H. Sadreddini, Mostafa Sheikhalishahi & Richard Wallace, Fatemeh Alimardani*

The general corpus contains all international news gathered from different news agencies. One demonstration of this comparison is shown in Figure 4. As we expected, when the input data is more comparable (i.e. sport related news), the efficiency of using the extracted data increases.

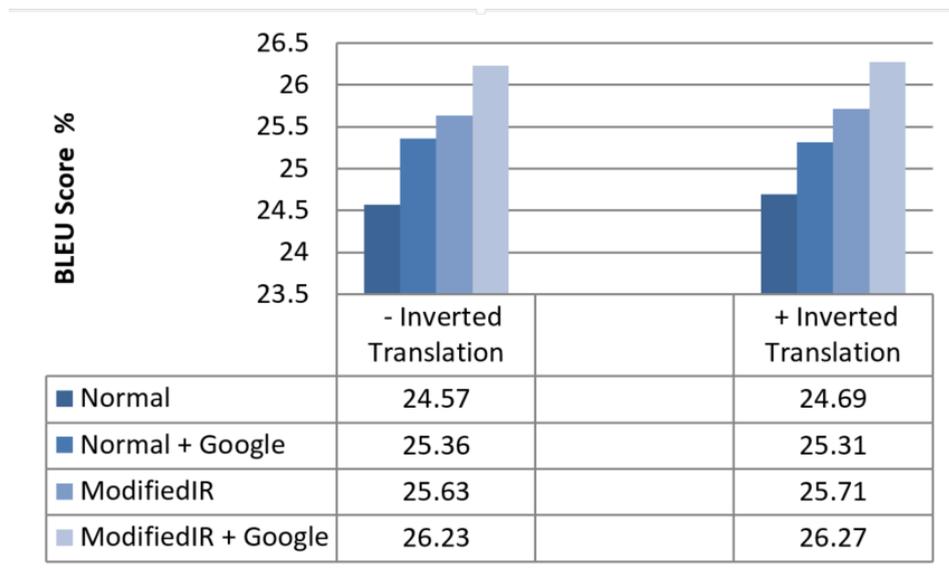

Figure 2: Effect of using Normalized Google Distance and invert translation with using general comparable corpus. The Normalized Google Distance is evaluated in both preliminary phase (+ Google) and IR system (ModifiedIR). All experiments have be done with and without inverted translation in filtering phase

Another evaluation performed in our experiments was to check the effect of different metrics in filtering phase. We applied all of the three different filters TER, TERp and WER individually. This evaluation is shown in Figure 5. Moreover, we used our approach to see the effect of these filters in several different cases. As can be seen in this figure, TER and TERp outperforms WER filter, and in most cases TER have better result in comparison with TERp. These results are very similar to which Abdul-Rauf and Schwenk (2011) indicated in their work.

To check the efficiency of our implemented approach in comparison with classic method introduced in Abdul-Rauf and Schwenk (2019/2011)'s work, the introduced small Persian-Italian parallel corpus is used to train the baseline SMT part of classic algorithm and its results are compared with ours. This comparison is shown in Figure 6. As can be seen in this figure, all experiments have been done with two different initial data (general news and international sport based news). The filtering scores TER was used in our implementation. In this experiment,

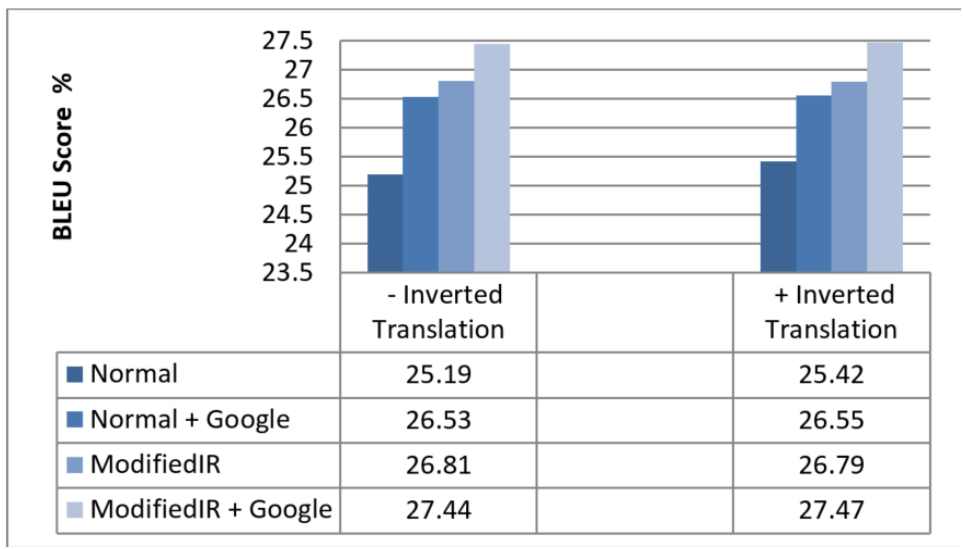

Figure 3: Effect of using Normalized Google Distance and invert translation with using international sport based (specific domain) comparable corpus. The Normalized Google Distance is evaluated in both primary phase (+ Google) and IR system (ModifiedIR). All experiments have be done with and without inverted translation in filtering phase

we used both of our normal and modified IR systems. In legends of this figure, "ModifiedIR" means we are using modified IR system which uses Google similarity metric and term "IT" means using inverted translation at filtering phase. However, these types of comparison are not quite fair when the size of initial parallel corpus used to build baseline system of classic version is not enough.

Finally, in order to see the effect of adding new extracted data in one real statistical machine translation system, Figure 7 shows the effect of adding different types of extracted sentences to our small existing Persian-Italian corpus. As can be seen there, extracted sentences could improve the initial machine translation system drastically. As we expected, the improvement is more when the comparable corpus (and consequently our test data) is a specific-domain corpus, with a higher comparability degree.

## 5 Conclusion

In this paper, we have introduced a new method to extract parallel sentences from comparable corpora. In our approach instead of simple translation and consequently comparison of source part to target part, a third language is used as the pivot. This model is suitable for those pairs of languages with small parallel cor-

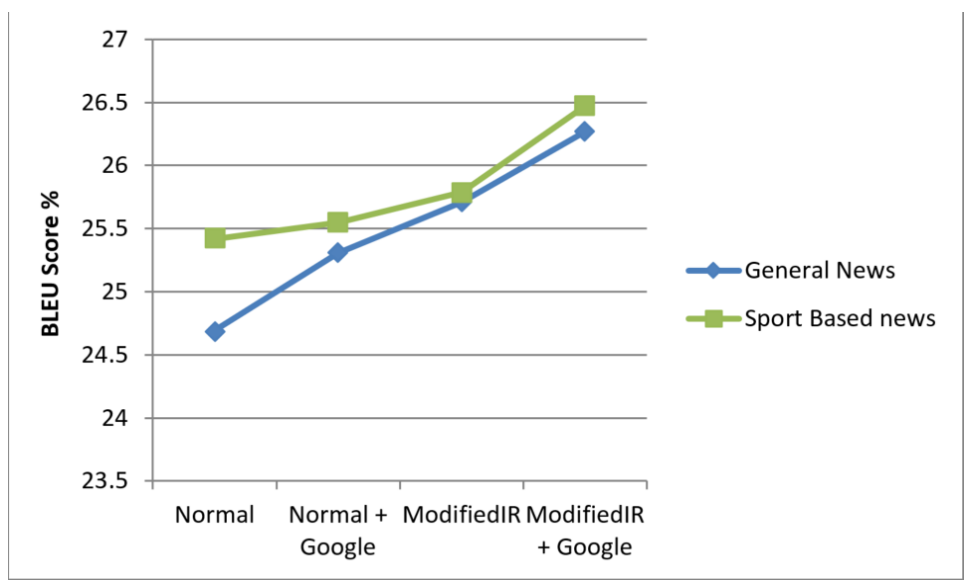

Figure 4: Comparison between different types of initial data. International news (general) and international sport news (domain based)

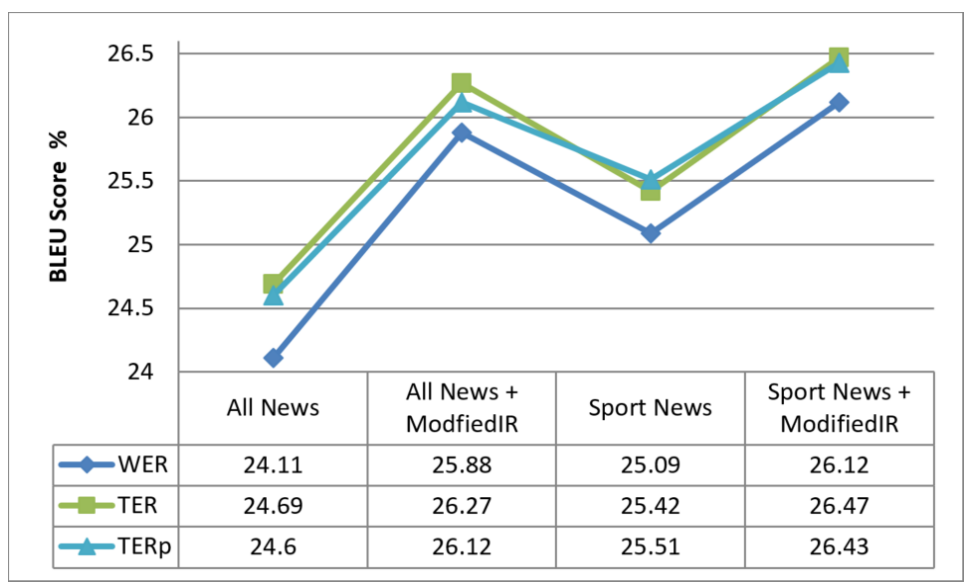

Figure 5: The effect of different filtering metric WER, TER and TERp on different comparable corpora (General News and international Sport based news) with and without Normalized Google Distance

pus and where there is a pivot language with sufficient amount of parallel corpus between it and both source and target languages. In our experiments the source, target and intermediate languages are Persian, Italian and English respectively. Considering that, there exist proper sizes of Persian-English and Italian-English

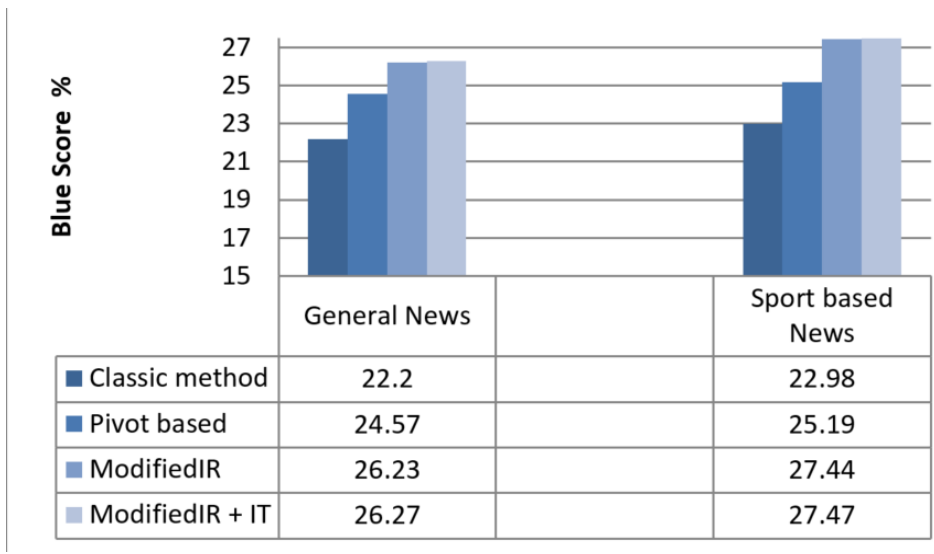

Figure 6: Comparison between different types of new approach with classic approach

parallel corpora, the initial SMTs were built on them. Based on this setup, the initial experimental results show the efficiency of our approach. Moreover, in order to improve the initial approach, we changed the adopted IR process of the classic algorithm which was introduced in (Abdul-Rauf and Schwenk 2011), and we added another similarity metric in different parts of the baseline system. This similarity metric is based on Normalized Google Distance (NGD). The extracted parallel corpus was improved by using this similarity metric in the primary phase (before our IR system) and inside the IR system. Moreover, using the similarity metric provides another threshold to prune and reduce the search space, and consequently enables us to increase the time window of our IR process. Thus the chance of finding a correct translation could be enhanced. In addition, we changed the filtering phase by using inverted translation of candidate sentences instead of the actual ones. While the size of our initial parallel Persian-Italian corpus was too small, this modification did not have a very big positive effect on the results. It is considerable that the effect of this new filtering phase could be evaluated with a larger size of Persian-Italian parallel corpus. Another issue which we considered was to see the effect of degree of comparability on the extracted sentences. Our experiment shows that, using texts which are more comparable (e.g. gathered news on a specific field) in comparison with texts with less comparability degree (e.g. general news) leads to better and more accurate extracted data. In order to check the effect of extracted sentences on a real statistical machine translation system, we added the new extracted corpus to our

*Ebrahim Ansari, M.H. Sadreddini, Mostafa Sheikhalishahi & Richard Wallace, Fatemeh Alimardani*

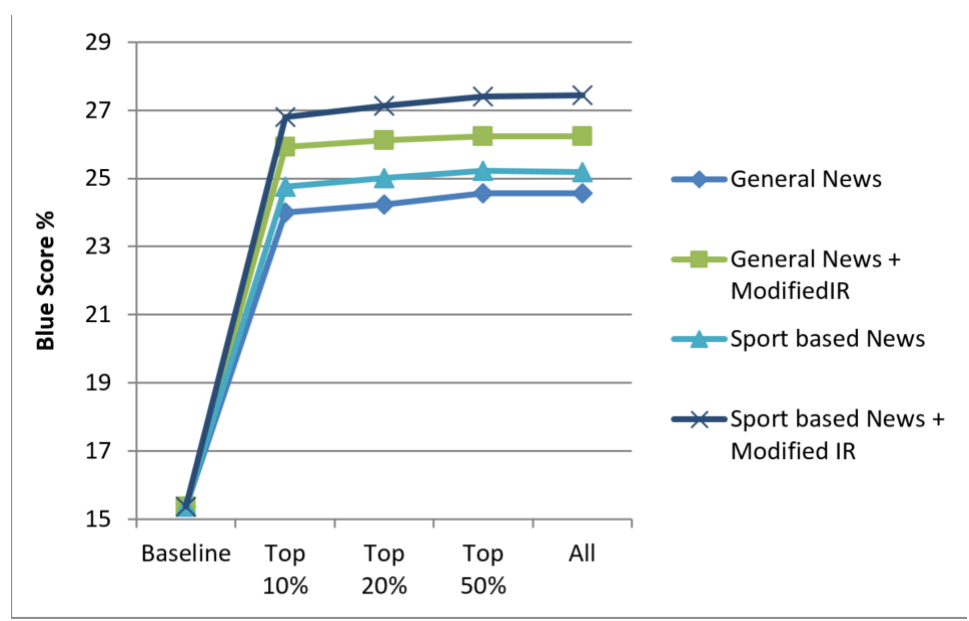

Figure 7: The effect of adding extracted data on real machine translation system. Four different sizes of extracted data are added to baseline machine translation system's corpus: Top 10% of extracted data, top 20% of extracted data, top 50% of extracted data and all the extracted data.

small Persian-Italian parallel corpora. Although the initial corpus was very small, the need for accurate parallel corpora to build an acceptable SMT is inevitable. The results show that adding extracted sentences from comparable corpus has a large effect on translation accuracy and improves our Persian-Italian statistical machine translation system.

# 6 Acknowledgment

The authors gratefully acknowledge the contributions and helps of Daniele Sartiano, Amir Onsori, S. M. H. Mirsadeghi, and Dr. M. N. Makhfif to this work.

# Name index

*Name index*